\documentclass{cvmp_short}
\usepackage{graphicx}

\begin{document}

\title{Real-Time Monocular 4D Face Reconstruction using the LSFM models}


\addauthor{Mohammad Rami Koujan}{}{1}
\addauthor{Nikolai Dochev}{}{1}
\addauthor{Anastasios Roussos$^{1,}$}{}{2}

\addinstitution{
Department of Computer Science,\\
 University of Exeter}
\addinstitution{
 Department of Computer Science,\\ Imperial College of London}
\maketitle


\noindent
\section{Introduction}
4D face reconstruction from a single camera is a challenging task, especially when it is required to be performed in real time. We demonstrate a system of our own implementation that solves this task accurately and runs in real time on a commodity laptop, using a webcam as the only input. Our system is interactive, allowing the user to freely move their head and show various expressions while standing in front of the camera. As a result, the put forward system both reconstructs and visualises the identity of the subject in the correct pose along with the acted facial expressions in real-time. The 4D reconstruction in our framework is based on the recently-released Large-Scale Facial Models (LSFM) \cite{LSFM1, LSFM2}, which are the largest-scale 3D Morphable Models of facial shapes ever constructed, based on a dataset of more than 10,000 facial identities from a wide range of gender, age and ethnicity combinations. This is the first real-time demo that gives users the opportunity to test in practice the capabilities of the recently-released Large-Scale Facial Models (LSFM) \cite{LSFM1, LSFM2}.

\section{Proposed Implementation}
The designed application relies on three main steps to generate the 4D facial reconstruction results:\\

\textbf{Facial landmarks detection}. Once a face gets detected, a set of facial landmarks are extracted from each input frame and used in later stages for estimating the 3D shape of that face. As our application is real-time, the landmarks extraction step needs to be fast. Accuracy and robustness of the landmarker are of paramount importance since the fitting approach relies solely on landmarks for reconstruction and the application does not necassitate any controlled environment. Toward these requirements, we adopt the method of Kazemi et al. \cite{landmarker} which offers a trade-off between the aspired landmarker's characteristics. When this landmarker fails, which might happen in case of occlusions or extreme poses, we switch to the more accurate (yet more computationally expensive) landmarker of Bulat et al. \cite{landmarker2}

\textbf{3D Morphable Model (3DMM)}. Our application makes use of \\
3DMMs fitting for the reconstruction task \cite{vetter}. As in [3], our adopted 3DMMs are derived from a linear combination of facial identity and expression variation:  
The identity part originates from the LSFM models (either global or bespoke) and the expression part originates from the blendshapes model of Facewarehouse \cite{facewarehouse}.  Fig. 1a shows the LSFM-global model mean shape ($\mu$) and first five principal components. The combination of both LSFM and Facewarehouse has first been proposed and tested in \cite{in-the-wild} and \cite{koujan2018combining}. Booth et al. \cite{in-the-wild} combine both LSFM and Facewarehouse to create a 3DMM for both identity and expression, which they use along with their suggested framework for fitting 3DMMs to images and videos captured under challenging environments (in-the-wild). We also make use of this combination in our implementation to inspect more its ability and promises for real-time 4D face reconstruction. Fig. 1c demonstrates such combination of identity and expression models.\\

\begin{figure}
\centering
\includegraphics[scale=0.3]{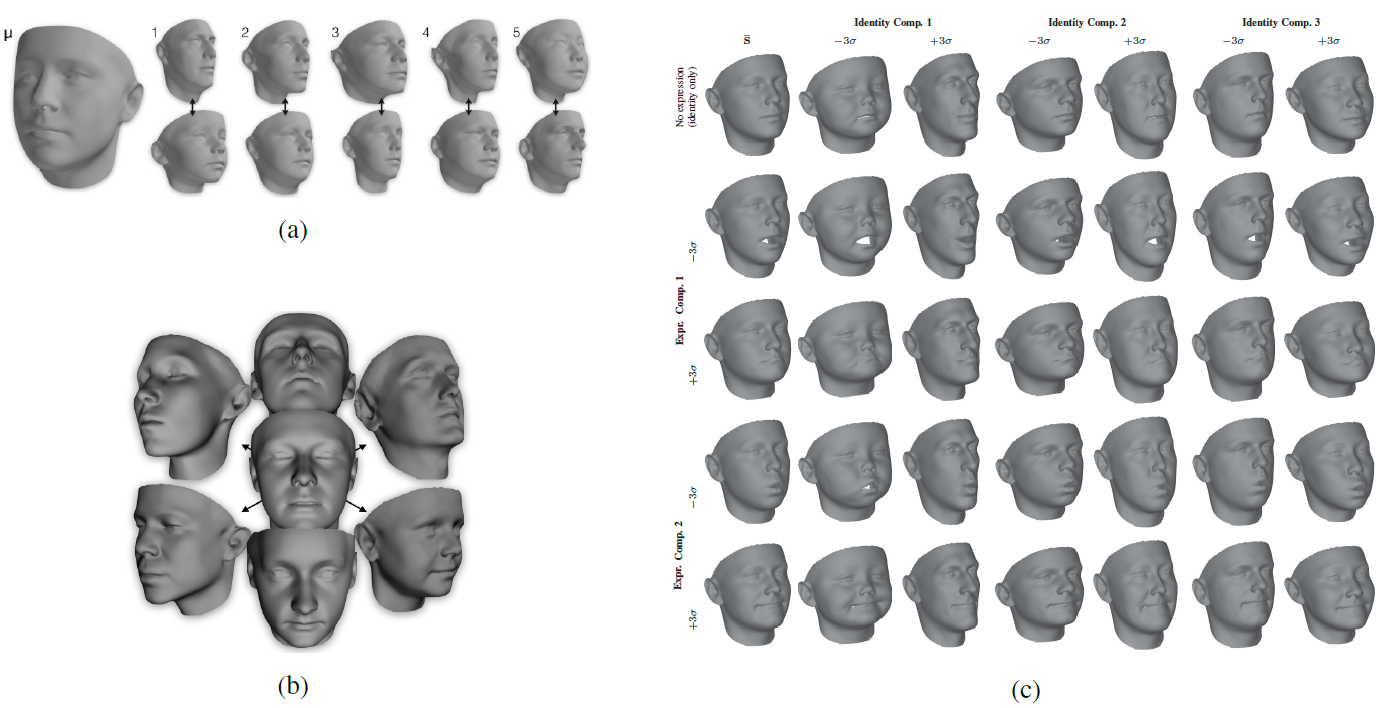}	
\caption{(a) Visualisation of the shape model of LSFM-global: mean shape ($\mu$) and first five principal components as additions and subtractions from the mean shape \cite{LSFM1}. (B) Exemplar instances generated from LSFM with distinct demographics. (c) Combined identity and expression model with fused mean faces and variations (-3 $\sigma$, 3 $\sigma$) along three principal components for identity, and two principal components for expression, with the first row showing only the identity part \cite{in-the-wild}.}
\label{fig:LSFM2}
\end{figure}

\textbf{Pose, Identity, and Expression Estimation}. To start with, we estimate the subject's face pose in each input frame based on an orthographic camera model. The idea is to use the extracted landmarks from the current frame along with their correspondences on the mean 3DMM face to estimate the pose, boiling down to a linear system of equations, which can be solved efficiently using Singular Value Decomposition (SVD). Estimating the identity and expression amounts to minimising a linear least squares problem similar to the one addressed in \cite{menpo-challenge}. To avoid any jitter effects, a temporal smoothing step is employed, using the average pose of three consecutive frames before visualising the result. This ensures that camera inherent noise and inacurracies associated with the landmarks will have minimal repercussions on the final result.

\subsection{Results}
\begin{figure}[h!]
\centering
\includegraphics[scale=.13]{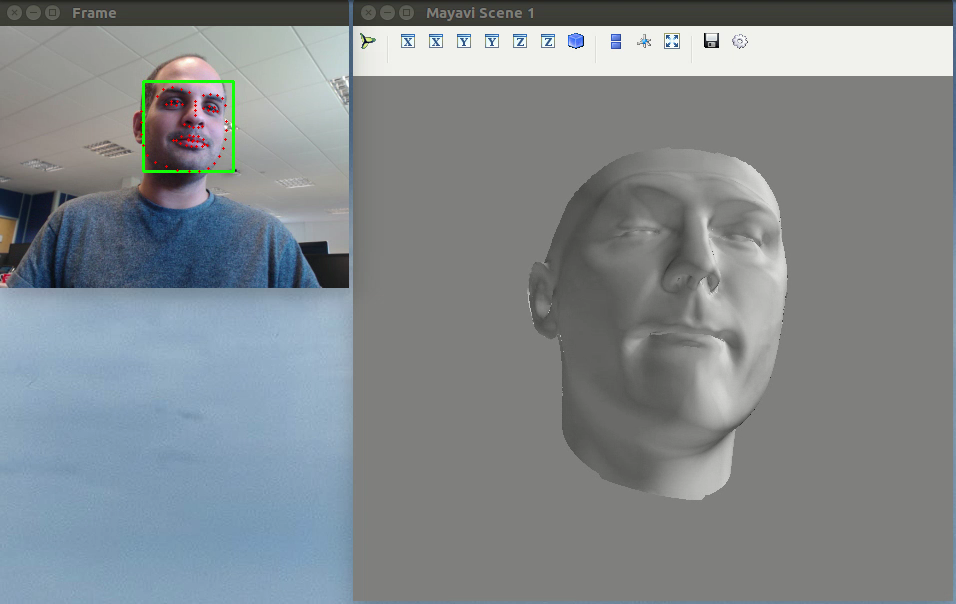}
\includegraphics[scale=.13]{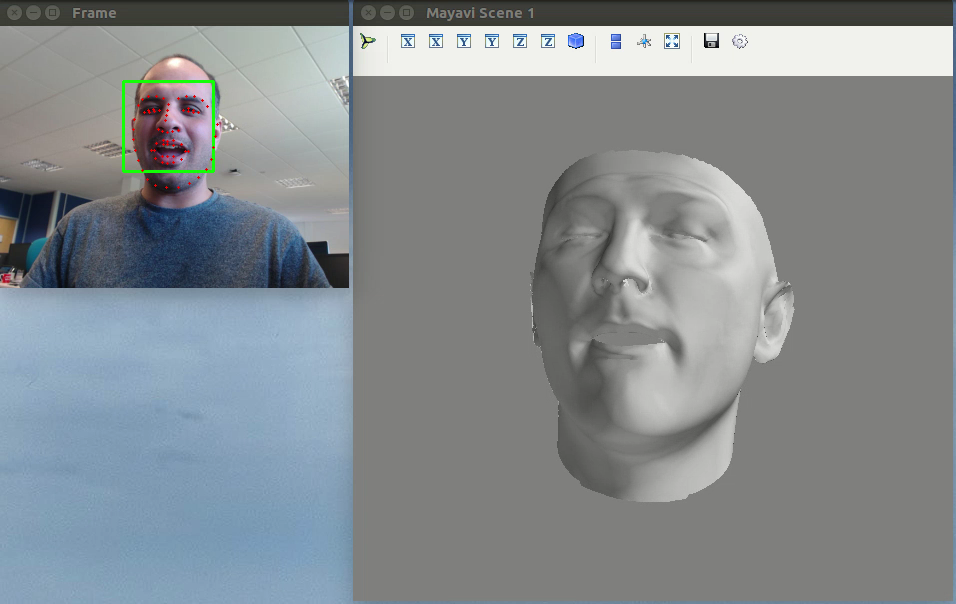}

\includegraphics[scale=.13]{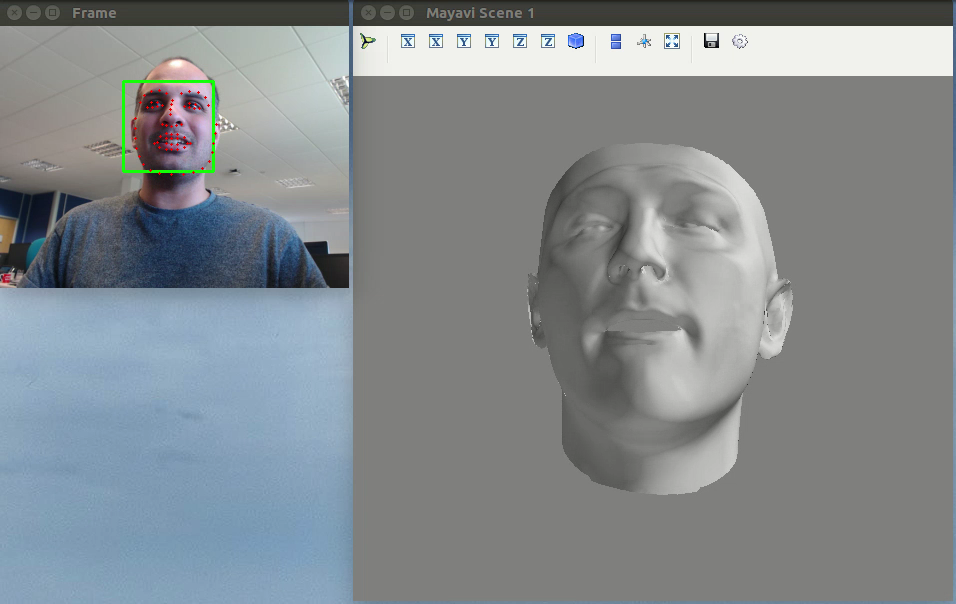}
\includegraphics[scale=.13]{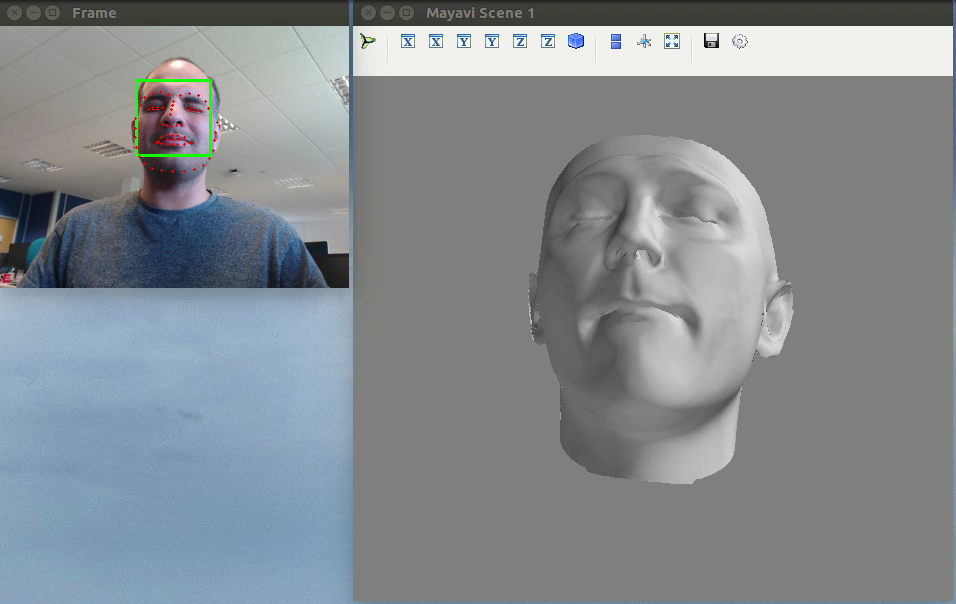}
\caption{Exemplar frames of reconstruction results of a subject showing different facial expressions in front of a camera.}
\label{fig:qualitative results}
\end{figure}

As described earlier, our application produces real-time results. Reaching such instantaneous  performance necessitated an efficient implementation tacking into account the intricate details to avoid any costly steps. Our application Graphical User Interface (GUI) offers the user a number of options to begin with. In addition to be used for normal reconstruction with/out showing the extracted landmarks with a bounding box around the detected face, it can visualise the reconstructed face with various exaggerated expressions, creating a caricature-like face. We provide the user with the ability to choose either the global or a bespoke LSFM model, among Black (all ages), Chinese (all ages) and White ethnic group, which is further clustered into four age groups: under 7 years old (White-under 7), 7-18 years old (White-7 to 18), 18-50 years old (White-18 to 50) and over 50 years old (White-over 50) \cite{LSFM1}. Fig. \ref{fig:qualitative results} presents the reconstruction results of a subject acting in front of a camera. 

\bibliography{egbib}

\end{document}